\documentclass{article}

\usepackage{arxiv}

\usepackage[utf8]{inputenc} 
\usepackage[T1]{fontenc}    
\usepackage{hyperref}       
\usepackage{url}            
\usepackage{booktabs}       
\usepackage{amsfonts}       
\usepackage{nicefrac}       
\usepackage{microtype}      
\usepackage{lipsum}
\usepackage{graphicx}
\graphicspath{ {./images/} }
\usepackage{subcaption}
\usepackage{amsmath}
\usepackage{amsthm}

\usepackage{float}

\title{Context-Sensitive Visualization of Deep Learning Natural Language Processing Models}

\author{
    Andrew Dunn\\
    Central Washington University\\
    Ellensburg, WA, USA\\
    \texttt{andrew.dunn@cwu.edu} \\
    \And
    Diana Inkpen\\
    University of Ottawa\\Ottawa, Canada\\
    dinkpen@uOttawa.ca\\
    \And
    R\u azvan Andonie \\
    Central Washington University\\
    Ellensburg, WA, USA\\
    and\\
    Transilvania University, Bra\c sov, Romania\\
    \texttt{razvan.andonie@cwu.edu} \\
}

\begin{document}

\maketitle

\begin{abstract}
The introduction of Transformer neural networks has changed the landscape of Natural Language Processing (NLP) during the last  years. So far, none  of  the visualization systems has yet managed to examine all the facets of the Transformers. This gave us the motivation of the current work. We propose a new NLP Transformer context-sensitive visualization method that leverages existing NLP tools to find the most significant groups of tokens (words) that have the greatest effect on the output, thus preserving some context from the original text. First, we use a sentence-level dependency parser to highlight promising word groups.  The dependency parser creates a tree of relationships between the words in the sentence. Next, we systematically remove adjacent and non-adjacent tuples of \emph{n} tokens from the input text, producing several new texts with those tokens missing. The resulting texts are then passed to a pre-trained BERT model. The  classification output is compared with that of the full text, and the difference in the activation strength is recorded. The modified texts that produce the largest difference in the target classification output neuron are selected, and the combination of removed words are then considered to be the most influential on the model's output. Finally, the most influential word combinations are visualized in a heatmap.
\end{abstract}

\section{Introduction}

The latest deep learning models are extremely complex, and can contain hundreds of millions of different trainable parameters. This high level of complexity makes it very difficult for humans to understand their inner workings. Visualization methods give us a window into these black box models, and can be beneficial for a number of reasons. For example, visualizations may provide a mechanism for debugging complex models by showing which parts of the input data the model is focusing on. This can be especially useful in classification models, because it may show that a model is focusing on parts of the data that are irrelevant to the predicted class.

Saliency maps are a popular tool for gaining insight into deep learning. In this case, saliency maps are typically depicted as heatmaps of neural layers, where "hotness" corresponds to regions that have a big impact on the model’s final decision. Grad-CAM is a state-of-the art technique for saliency map visualization \cite{Selvaraju2019} using the gradient information obtained from backpropagating the error signal from the loss function with respect to a specific feature map, at any layer of the network. 

While the idea of using visualization methods on image processing models makes intuitive sense, in recent years there has been a growing interest in visualizing NLP models. However, with NLP models visualization is not as straight forward due to working with text. Current solutions to visualizing NLP models involve heatmaps similar to Grad-CAM, and visualizing connections between tokens in models that utilize attention mechanisms. These solutions typically look at the importance of individual tokens on the model output. Before being passed to an NLP model, the text must be broken up into a series of tokens that are then transformed into numerical vectors. Each token typically represents a single word or character in the input text. The goal of these NLP visualization methods is to highlight the most significant parts of the text that have the greatest impact on the model output. While effective, current methods look at each separate token independently, thus lacking context from the entire text. We believe the combination and order in which the words appear is significant and that they influence an NLP model’s interpretation of the text.

Typically, for tasks such as machine translation, the output is in textual form, and an encoder-decoder mechanism is used. The weakness of this approach is that the encoder produces a finite-length vector, without regards to weather or not any of the inputs are more important (or less important). A solution to this problem is the \emph{attention} mechanism. Several variants of this mechanism have been introduced: convolutional, intra-temporal, gated, and self-attention \cite{Otter2019}. Self-attention has recently become very popular in the state-of-the art model called \emph{Transformer}, and provides attention to words in the same sentence. 

Visualization is considered a preliminary step for interpreting or explaining a trained machine learning model. The terms interpretability and explainability are often used interchangeably. However, it is generally accepted that there is a difference between them. Interpretability in a machine learning model means that the model behaves in a consistent and predictable manner, so that changes to model parameters result in expected changes in the model's output. Explainability refers to the process of providing \emph{post hoc} explanations for the behavior of black box models \cite{Marc2020}. In this paper, we are primarily concerned with the explainability of Deep Neural Network (DNN) models. 

We propose a new NLP Transformer context-sensitive visualization method that leverages existing NLP tools to find the most significant groups of tokens (words) that have the greatest effect on the output, thus preserving some context from the original text. First, we use a sentence-level dependency parser to highlight promising word groups.  The dependency parser creates a tree of relationships between the words in the sentence. Next, we apply a leave-$n$-out (LNO) process: we systematically remove adjacent and non-adjacent tuples of \emph{n} tokens from the input text, producing several new texts with those tokens missing. The resulting texts are then passed to a BERT-based model. The  classification output is compared with that of the full text, and the difference in the activation strength is recorded. The modified texts that produce the largest difference in the target classification output neuron are selected, and the combination of removed words are then considered to be the most influential on the model's output. In the visualization stage, these most influential word combinations are highlighted by a heatmap.

The rest of the paper is structured as follows. Section \ref{related} introduces basic concepts and describes related work.  Section \ref{proposed} describes our novel method. Experimental results are detailed and discussed in Section \ref{experiments}. Finally, Section \ref{conclusion} presents conclusions and open problems.

\section{Related Work: Visualization of Transformer-Based Deep Networks} \label{related}

One of the first papers introducing visualizations of neural NLP models is \cite{Li2016}. Four strategies for visualizing neural models for NLP were described, all inspired from similar work in computer vision. However, NLP visualization is more complex than visualization in image recognition models. Recently, more specialized techniques were developed for NLP visualization, the most popular one being Transformers-based deep networks  \cite{Brasoveanu2020}.  

The introduction of Transformer neural networks has changed the landscape of NLP during the last years. The Transformer  combines an encoder, decoder, and a self-attention mechanism to create a very effective model that is able to infer connections between inputs in sequential data, similar to recurrent neural networks like the popular long short-term memory (LSTM) models \cite{DBLP:conf/nips/VaswaniSPUJGKP17}. This allows the Transformer model to extract complex features from the input data, and has proven to be very effective in solving NLP problems.

BERT \cite{DBLP:conf/naacl/DevlinCLT19} is a state-of-the-art NLP model developed by Google in 2018 that has seen many successes in common NLP tasks such as classification, sentence prediction, named entity recognition, and more. BERT stands for Bidirectional Encoder Representations from Transformers, and is built using a stack of Transformer encoders. BERT is a very large and deep network containing 110 million trainable parameters (BERT-base) or 345 million parameters (BERT-large), which makes training BERT from scratch very time consuming. Due to the large number of parameters, explaining the individual components or behavior of the model is very difficult.

The success of BERT has lead to researchers to create several similar language models based on Transformers, such as RoBERTa \cite{DBLP:journals/corr/abs-1907-11692}, DistilBERT \cite{DBLP:journals/corr/abs-1910-01108}, and XLNet \cite{DBLP:conf/nips/YangDYCSL19} that are aimed at either improving the accuracy of the model, or reducing the training time and the complexity of the model.

GPT-2 is another very large language model developed by OpenAI in 2019 \cite{Radford2019}. Like BERT, GPT-2 is based on Transformers and was designed to come in multiple sizes. GPT-2 Large contains 762 million trainable parameters, making it considerably larger than BERT-large. Where BERT is built using a stack of Transformer encoders, GPT-2 is made up from a stack of Transformer decoders. In addition, GPT-2 uses masked self-attention, which limits attention weights between the current and previous tokens in the sequence. In contrast, BERT uses unmasked self-attention, which allows attention weights to be generated between all tokens in the text, regardless of their position. In 2020, OpenAI released GPT-3 \cite{Brown2020}, a massive language model composed of 175 billion trainable parameters. At the time of completion, GPT-3 was the largest DNN language model ever created. In certain applications, GPT-3 has been able to generate text that indistinguishable from human written text. As of the writing of this paper, GPT-3 is not  publicly available for download.

Explaining the information processing flow and results in a Transformer is difficult because of its complexity. A convenient and very actual approach is visualization. Visualization enables some degree of explainability. A first survey on visualization techniques for Transformers is \cite{Brasoveanu2020}. 

Visualizing complex models like those built with Transformers, can sometimes be accomplished by using model-agnostic tools specifically built for benchmarking or hyperparameter scoring, such as \emph{Weights and Biases}\footnote{https://www.wandb.com/} - which provides the largest sets of visualization and customization capabilities. Perhaps more interesting are the visualizations  specifically built around Transformers, either for the purpose of explaining the models (like ExBERT \cite{DBLP:journals/corr/abs-1910-05276}), or for explaining certain model-specific attributes (like embeddings or attention maps \cite{DBLP:conf/acl/Vig19}). Details are provided in \cite{Brasoveanu2020}, including a list of recent articles focused on explaining Transformer topics through visualizations.

In addition to Transformer visualizations, there are several specific BERT visualization approaches \cite{Brasoveanu2020} focused on  the role of embeddings and relational dependencies within the Transformer learning phase \cite{DBLP:conf/nips/ReifYWVCPK19}, the role of attention during pre-training or training \cite{DBLP:conf/iclr/SuZCLLWD20, DBLP:conf/acl/Vig19}, and the importance of various linguistic phenomena encoded in its language model \cite{DBLP:journals/corr/abs-1906-04341}. 

A general goal would be to visualize the entire lifecycle of a Transformer models. However, none of the current visualization systems (considered in \cite{Brasoveanu2020}) has yet managed to examine all the facets of the Transformers. This is due to the fact that this area is relatively new and there is no consensus on what needs to be visualized. The visualization of NLP neural models is still rather young, and we believe that our proposed method brings a new and interesting insight.

Recently, Han \emph{et al.} \cite{Han2020} mapped saliency over NLP  inputs to explain predictions via influential training examples using a BERT-based model. They found that in a lexicon-driven sentiment analysis task, saliency maps and influence functions are largely consistent with each other. 

However, Han \emph{et al.}'s \cite{Han2020} results were not consistent for the more complex task of Natural Language Inference, a classification problem that concerns the relationship between a premise sentence and a hypothesis sentence. An explanation could be that saliency maps focus on the input, and may neglect to explain how the model makes decisions. Some saliency methods are incapable of supporting tasks that require explanations that are faithful to the model or the data generating process \cite{Adebayo2018}.

Very recently, Yann LeCun's team from Facebook AI Research, UC Berkeley and New York University applied dictionary learning techniques to provide detailed visualizations of transformer representations and insights into the semantic structures \cite{Yun2021}. Their hypothesis was that contextualized word embedding can serve as a sparse linear superposition of Transformer factors. They demonstrated that the hierarchical semantic structures captured by the Transformer factors, e.g.,  word-level polysemy disambiguation, sentence-level pattern formation, and long-range dependency. They have created an interactive website\footnote{https://transformervis.github.io/transformervis/} where users can gain additional insights into transformer models by visualizing their latent space. They used a pretrained BERT model for their evaluation experiments. Transformer factors were divided into three different levels: low level (word-level), mid level (sentence-level) and high level (long range patterns).

To the extend of our knowledge, the closest existent result to our contribution is \cite{Yun2021}. Our hypothesis is also that contextualized word embedding can be valuable for visualization in transformer models. Our work was performed independent of \cite{Yun2021}, and we had no time to experimentally compare the two methods.

\section{Proposed Method: Context-sensitive visualization} \label{proposed}

The success of very large language models, such as BERT and GPT-2, has created an interest in visualizing these complex models so that we can better understand how they work and explain their behavior. With these goals in mind, we propose a new method for visualizing NLP classification models. 

Previous visualization solutions, like leave-one-out (LOO) and saliency maps, look at each input token individually. Since new language models such as BERT are capable of understanding complex phrases and relationships between words, we believe new visualization methods that preserve some context from the original text will improve the explainability of these models. 

The LOO process is a feature erasure technique that systematically identifies and removes a single feature from the input data. In an NLP model, a feature is the vector representing a single token in the input text, typically a word or character, although character n-grams and subwords may also be selected as tokens. The modified data is then passed through the model and the difference in output strength is recorded. This process is repeated for every feature. The token influence is calculated by subtracting the output strength of the modified inputs from the output strength of the original unmodified input. The output strength of the removed features are then ranked in order of their influence score, and all tokens that had a positive difference in output value when removed are considered as the set of most influential features on the model's output. 

Based on LOO, the LNO process systematically removes adjacent and non-adjacent tuples of \emph{n} tokens from the input text, producing several new texts with those tokens missing. Where LOO removes a single token at a time, LNO removes multiple tokens at each step. The resulting texts are then passed to the model and their classification output is compared with that of the full text, and the difference in the activation strength is recorded. The modified texts that produce the largest difference in the target classification output neuron are selected, and the combination of removed words are then considered to be the most influential on the model's output. In our experiments, we elected to use word pairs, and thus use tuples of size two. Tuples of more than two tokens could be also used, and may provide different results. However, the computational overhead may increase in this case exponentially and the benefits are questionable.

Our approach looks at combinations of words and phrases to discover their effect on the model's output, producing a visualization that is more context sensitive to the original text. Our method leverages existing NLP tools for dependency parsing combined with a model-agnostic LNO token pre-processing approach to identify and highlight contextual groups of tokens that have the largest perceived effect on the model's classification output.

\begin{figure}
    \centering
    \includegraphics[width=0.6\textwidth]{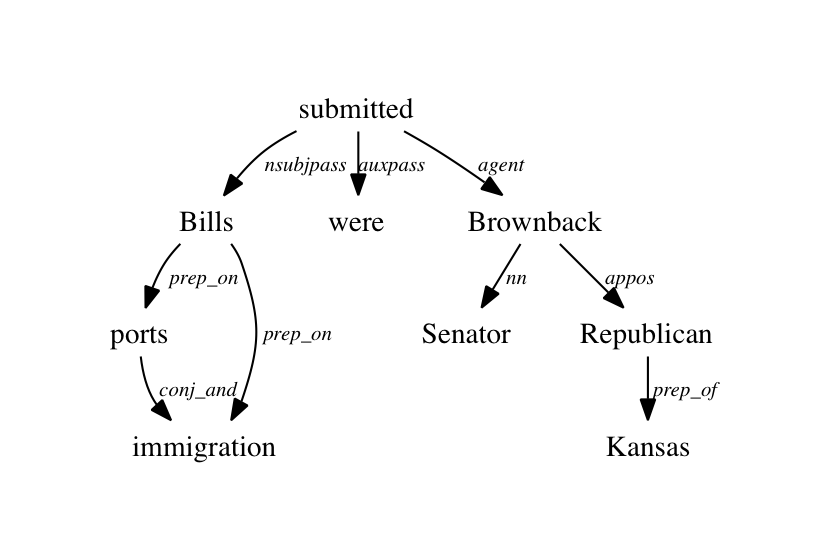}
    \caption{Dependency parser example output \cite{Danqi2014} \cite{stanforddepend}}
    \label{fig:depend}
\end{figure}

Since the LNO processing step is computationally expensive and the number of possible word combinations explodes exponentially, we use a sentence-level dependency parser to highlight promising word groups. This step greatly reduces the complexity of processing a sentence. The dependency parser creates a tree of relationships between the words in the sentence. By only selecting groups of words that are connected via a branch in the dependency tree, we are able to greatly reduce the search space for the algorithm. Figure \ref{fig:depend} shows a visual example of a dependency tree generated by the Stanford Dependencies library, created at the Stanford University\footnote{https://nlp.stanford.edu/software/nndep.html}.  The original sentence that was given to the dependency parser as follows: \textit{Bills on ports and immigration were submitted by Senator Brownback, Republican of Kansas}.

\begin{figure}
    \centering
    \includegraphics[width=0.5\textwidth]{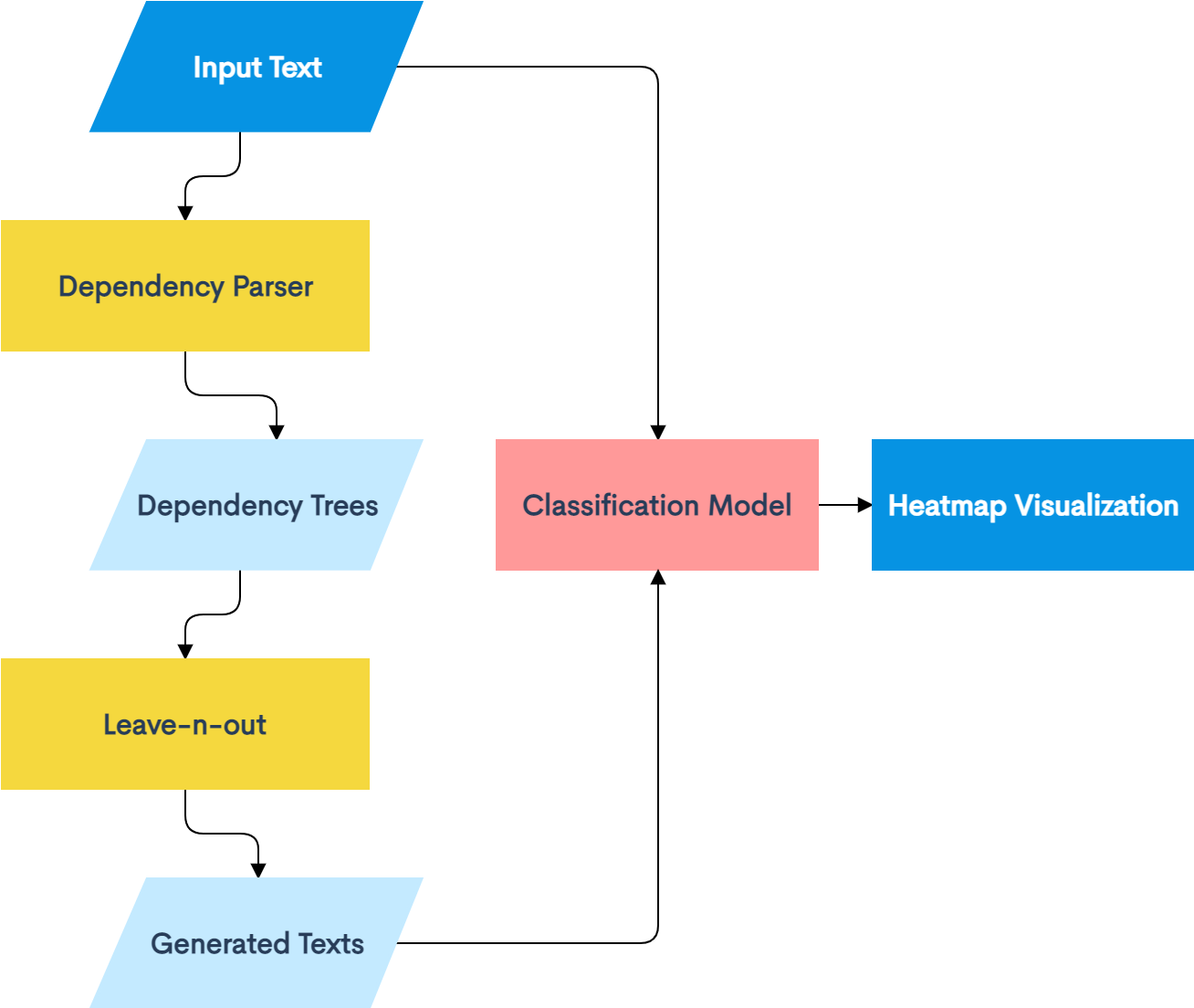}
    \caption{LNO process flowchart}
    \label{fig:lno-flowchart}
\end{figure}

Using the previously described processes, the steps required to visualize a text using our method are as follows. First, an input text is passed through the classification model and the class prediction and output strength is recorded as the baseline. Next, each sentence in the text is passed through a dependency parser which produces a tree structure highlighting dependency relations between the various words in each sentence. Using these dependency trees, new texts are generated with word pairs removed. Word pairs are limited to those that are connected via a branch in the dependency tree. Each generated text has only one word pair removed at a time, and words are unique to their position in the sentence. After all new texts are generated, we proceed to the LNO process. All generated texts are passed through the model and their class prediction and output strengths are recorded. Once all texts have been processed, the output strength of the predicted class neuron for each generated text is then compared to output strength of the original baseline text. The heatmap is then generated by first sorting all word pairs by their output strength difference to that of the original text. Word pairs that did not contribute to the class prediction are culled. If an individual word appears in more than one pair, the maximum output strength for that word out of all pairs is selected. Finally, the output strengths for all remaining words is normalized using a linear scale. Figure \ref{fig:lno-flowchart} contains a simplified flowchart detailing the order of operations for our method.

\section{Experiments} \label{experiments}

For our experiments, we used the IMDB Large Movie Review sentiment analysis dataset\footnote{https://ai.stanford.edu/~amaas/data/sentiment/} from Stanford University, which contains 25,000 training and 25,000 testing reviews labelled with two categories: positive reviews and negative reviews. We used 20\% of the training set as a separate validation dataset. As mentioned, the test set is separate from the training and validation sets.

All code is written in Python 3.8, and utilizes the Tensorflow version of the Transformers library\footnote{https://huggingface.co/transformers/}. A dropout layer and a fully connected layer are appended to the pretrained BERT-base model to create our sentiment analysis classification model. Text tokenization and dependency parsing is executed using the spaCy NLP library\footnote{https://spacy.io/}. The visualization images are generated using the Matplotlib library\footnote{https://matplotlib.org/}.

In the following, we will present some experimental results and heatmap visualizations for both the LOO and LNO methods.  

\begin{figure}[H]
    \centering
    \includegraphics[width=0.5\textwidth]{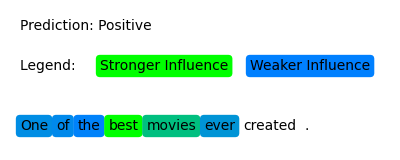}
    \caption{Example LOO heatmap output - positive review}
    \label{fig:loo-ex1}
\end{figure}

\begin{figure}[H]
    \centering
    \includegraphics[width=0.5\textwidth]{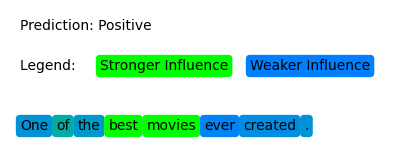}
    \caption{Example LNO heatmap output - positive review}
    \label{fig:lno-ex1}
\end{figure}

Figures \ref{fig:loo-ex1} and \ref{fig:lno-ex1} show a short example text that represents a positively-classified review. In Figure \ref{fig:loo-ex1}, the LOO process identified the word "best" as the most influential to the classification. We can see in Figure \ref{fig:lno-ex1} that more of the sentence context is highlighted, with the word pair "best movies" most prominently highlighted.

\begin{figure}[H]
    \centering
    \includegraphics[width=0.5\textwidth]{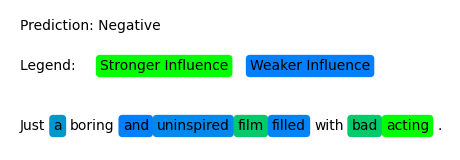}
    \caption{Example LOO heatmap output - negative review}
    \label{fig:loo-ex2}
\end{figure}

\begin{figure}[H]
    \centering
    \includegraphics[width=0.5\textwidth]{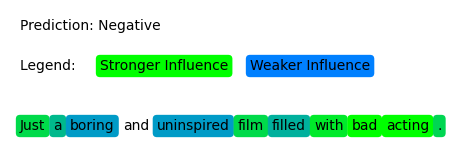}
    \caption{Example LNO heatmap output - negative review}
    \label{fig:lno-ex2}
\end{figure}

Figures \ref{fig:loo-ex2} and \ref{fig:lno-ex2} show another short example text, this time a negatively-classified review. Figure \ref{fig:loo-ex2} shows the visualization result from the LOO process, and the word "acting" is most prominently highlighted, with "film" and "bad" also showing a stronger influence. In Figure \ref{fig:lno-ex2}, we can see the results from the LNO process, and overall the influence strength of individual words is increased across the board, with the word pair "bad acting" highlighted as the most influential. Also note that the word "boring" is not considered in the LOO process, but it is highlighted in the LNO process.

\begin{figure}[H]
    \centering
    \includegraphics[width=0.5\textwidth]{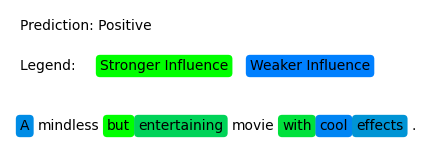}
    \caption{Example LOO heatmap output - mixed positive review}
    \label{fig:loo-ex4}
\end{figure}

\begin{figure}[H]
    \centering
    \includegraphics[width=0.5\textwidth]{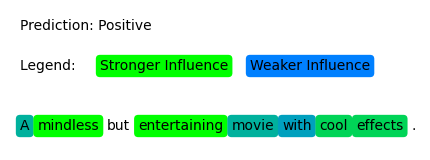}
    \caption{Example LNO heatmap output - mixed positive review}
    \label{fig:lno-ex4}
\end{figure}

Figures \ref{fig:loo-ex4} and \ref{fig:lno-ex4} show a mixed positively-classified review. Again, Figure \ref{fig:loo-ex4} shows the resulting visualization from the LOO process. Note that the word "mindless" is not highlighted, but the word "but" is. Interestingly, Figure \ref{fig:lno-ex4} shows that heatmap result from LNO has reversed for these two words, with "mindless" highlighted with a stronger influence on the classification and "but" being not highlighted. Also note that the LOO result gives the word "with" a stronger influence than the words "cool effects", which is contrary to what we might expect. In contrast, the LNO heatmap gives more importance to the "cool effects", which is more inline with our expectations.

\begin{figure}[H]
    \centering
    \includegraphics[width=0.5\textwidth]{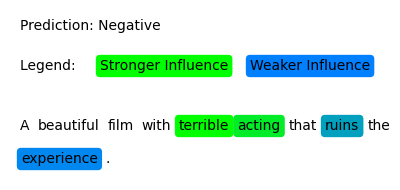}
    \caption{Example LOO heatmap output - mixed negative review}
    \label{fig:loo-ex3}
\end{figure}

\begin{figure}[H]
    \centering
    \includegraphics[width=0.5\textwidth]{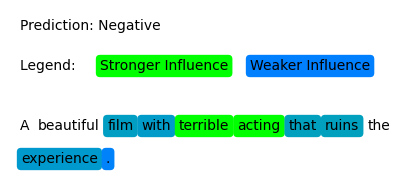}
    \caption{Example LNO heatmap output - mixed negative review}
    \label{fig:lno-ex3}
\end{figure}

Figures \ref{fig:loo-ex3} and \ref{fig:lno-ex3} show another mixed review, but this time negatively-classified. In both the LOO and LNO heatmaps we can observe that the word pair "terrible acting" has the strongest influence on the classification. Since this was a negative sentiment prediction, the word "beautiful" appears to have no influence on the classification shown in both heatmaps. As hoped, although the results between the two approaches is similar, the LNO process picks up on more of the review context than LOO, with more of the relevant text highlighted.

\section{Conclusion}\label{conclusion}

Given that NLP tasks require some special considerations due to the nature of how languages are represented and understood, visualization tools that are tailored specifically for NLP models may be preferred over more traditional methods, such as saliency maps based on gradients. The knowledge contained within a text goes beyond the meaning of the its individual words. The relationships between words and their positions within the sentence are necessary for a full understanding of the text. Thus, we believe a visualization method that retains some of the contextual meaning between the different words in the input text will produce improved results that are more inline with our expectations. Improved visualizations increase our understanding of the model, and help us design better models that perform closer to the human level of understanding for these problems. With the method presented in this paper, we have attempted to create a context-sensitive visualization method for deep NLP classification models. This more context sensitive approach leads to heatmaps that include more of the relevant information pertaining to the classification, as well as more accurately highlighting the most important words from the text.

The terms attention and explanation are used frequently in the context of visualization of deep networks, for instance when using saliency (or heatmaps) as visualization tools. While attention modules may yield improved performance on NLP tasks, their ability to provide meaningful explanations for model predictions is questionable. However, according to \cite{Jain2019}, attention weights over non-recurrent encoders exhibit relatively strong correlations, on average, with LOO feature importance scores and gradient-based measures. Since we use LNO and gradient-based techniques, we may expect to have some correlation between attention weights and feature importance scores for the top features. We leave this aspect as a direction of investigation in future work.

\bibliographystyle{IEEEtran}
\bibliography{main}

\end{document}